\documentclass[letterpaper]{article} % DO NOT CHANGE THIS
\usepackage{aaai25}  % DO NOT CHANGE THIS
\usepackage{times}  % DO NOT CHANGE THIS
\usepackage{helvet}  % DO NOT CHANGE THIS
\usepackage{courier}  % DO NOT CHANGE THIS
\usepackage[hyphens]{url}  % DO NOT CHANGE THIS
\usepackage{graphicx} % DO NOT CHANGE THIS
\usepackage{natbib}  % DO NOT CHANGE THIS AND DO NOT ADD ANY OPTIONS TO IT
\usepackage{caption} % DO NOT CHANGE THIS AND DO NOT ADD ANY OPTIONS TO IT
\usepackage{algorithm}
\usepackage{algorithmic}
\usepackage{svg}
\usepackage{multirow}
\usepackage[countmax]{subfloat}
\usepackage{subfig}
\usepackage{newfloat}
\usepackage{listings}
\DeclareCaptionStyle{ruled}{labelfont=normalfont,labelsep=colon,strut=off} % DO NOT CHANGE THIS
\floatstyle{ruled}
\newfloat{listing}{tb}{lst}{}
\floatname{listing}{Listing}
\title{LoRA Unlearns More and Retains More (Student Abstract)}
\author{
   Atharv Mittal }

\affiliations{
    Vision and Language Group, Indian Institute of Technology Roorkee, Roorkee, Uttarakhand, India - 247667 \\
   atharv\_m@mfs.iitr.ac.in   
}

\begin{document}

\maketitle

\begin{abstract}
Due to increasing privacy regulations and regulatory compliance, Machine Unlearning (MU) has become essential. The goal of unlearning is to remove information related to a specific class from a model. Traditional approaches achieve exact unlearning by retraining the model on the remaining dataset, but incur high computational costs. This has driven the development of more efficient unlearning techniques, including model sparsification techniques, which boost computational efficiency, but degrade the model's performance on the remaining classes.
To mitigate these issues, we propose a novel method, PruneLoRA which introduces a new MU paradigm, termed prune first, then adapt, then unlearn. LoRA \cite {hu2022lora} reduces the need for large-scale parameter updates by applying low-rank updates to the model. We leverage LoRA to selectively modify a subset of the pruned model’s parameters, thereby reducing the computational cost, memory requirements and improving the model's ability to retain performance on the remaining classes. Experimental Results across various metrics showcase that our method outperforms other approximate MU methods and bridges the gap between exact and approximate unlearning.
Our code is available at \url{https://github.com/vlgiitr/LoRA-Unlearn}.

\end{abstract}

\section{Introduction}{
The process of removing specific data points or classes from trained machine learning models is known as machine unlearning. Its importance has intensified due to growing privacy concerns and the need to comply with evolving regulations, which enable users to request the removal of their personal data from models as part of the “right to be forgotten” in General Data Protection Regulation (GDPR). \newline
Machine unlearning techniques can be classified into two broad categories: exact and approximate
unlearning. The exact approach to machine unlearning typically involves retraining the entire model on a modified dataset, excluding the data to be forgotten. While this method guarantees the removal of the influence of a data instance from a model, it is highly computationally intensive for larger models.
Approximate unlearning focuses on reducing the influence of targeted data points through efficient parameter updates. However, these methods often struggle to balance unlearning effectiveness with performance and computational efficiency. One common approximate unlearning method is simple fine-tuning (FT), which fine-tunes the pre-trained model on the remaining dataset for a few training epochs, but presents its own challenges. When a model is fine-tuned to forget specific information, it often suffers from catastrophic forgetting, i.e it loses the ability to perform well on previously learned tasks, thus degrading its performance on the remaining classes. Fine-tuning can also be computationally expensive on very large models. 
 \newline
To address these limitations, \cite{modelsparsity2024} explored model sparsification techniques, where they selectively removed specific weights or neurons within the model, rather than updating the entire network before fine-tuning. By focusing on a subset of the model's parameters, they reduced overfitting and computational cost. However, despite offering improvements over standard fine-tuning, these methods still encounter challenges in effectively balancing the trade-offs between unlearning efficiency, computational cost, and maintaining overall model performance.
Low-Rank Adaptation (LoRA), offers a solution that builds upon the principles of model sparsity while addressing its limitations by updating only a small subset of model parameters through low-rank matrix decomposition. Since \cite{biderman2024lora} shows that in the context of LLMs, LoRA provides a form of regularization that mitigates “forgetting” of the source domain, through rigorous experimentation, we prove using LoRA to update model parameters preserves model performance and lowers computational costs exponentially. 
}

\begin{figure*}[!t]
\centering
\resizebox{\textwidth}{!}{%
\begin{tabular}{|c|cc|cc|cc|cc|c|c|}
\hline
\multirow{2}{*}{\textbf{Model}} & \multicolumn{2}{c|}{\textbf{UA}} & \multicolumn{2}{c|}{\textbf{MIA-Efficacy}} & \multicolumn{2}{c|}{\textbf{RA}} & \multicolumn{2}{c|}{\textbf{TA}} & \textbf{RTE} & \textbf{GPU} \\ \cline{2-11} 
& \textbf{5 Epochs} & \textbf{10 Epochs} & \textbf{5 Epochs} & \textbf{10 Epochs} & \textbf{5 Epochs} & \textbf{10 Epochs} & \textbf{5 Epochs} & \textbf{10 Epochs} &\textbf{(secs/epoch)}  & \textbf{GB}  \\ \hline
\multicolumn{11}{|c|}{\textbf{ResNet-50}} \\ \hline
\textbf{Retrain}             & 100.00    & 100.00   &100.00   & 100.00 & 98.02   & 98.02   & 96.70 & 96.70 &- &- \\ 
\textbf{Finetune}             & 100.00    & 100.00   &100.00   & 100.00 & 92.52   & 96.27   & 88.29 & 91.44 &137 &6.9\\ 
\textbf{Pruned}                   & 100.00    & 100.00    & 100.00    & 100.00    & 94.92    & 95.68  & 90.79 & 90.72 &137 &4.2\\ 
\textbf{LoRA}                   & 97.22    & 100.00    & 100.00    & 100.00    & 96.90   & 97.19 & 95.03 & 93.49 &122 & 5.8 \\ 
\textbf{Pruned LoRA}                 & 99.78    & 99.98    & 97.68    & 97.89    & \textbf{97.96}  & \textbf{98.00}  & \textbf{95.18} & \textbf{95.41} &122 &5.5 \\ \hline

\multicolumn{11}{|c|}{\textbf{ViT}} \\ \hline
\textbf{Retrain}             & 100.00    & 100.00   &100.00   & 100.00 & 96.90   &  96.90  & 84.92 & 84.92 &- &-\\ 
\textbf{Finetune}             & 100.00    & 100.00   & 100.00  & 100.00 & 87.64   & 86.80   & 79.79 & 79.94 & 132 &3.2\\ 
\textbf{Pruned}                   & 100.00 & 100.00  & 100.00 & 100.00    & 84.76    & 86.14 & 78.81 & 79.53 & 132 &3.2\\ 
\textbf{LoRA}                 & 87.66    & 95.58    & 98.14    & 99.58    & \textbf{97.72}    & \textbf{97.81} & 85.33 & 85.16 &48 &0.7 \\ 
\textbf{Pruned LoRA}                   & 100.00    & 100.00    & 100.00    & 100.00   & 97.39    & 97.63 & \textbf{85.53} &\textbf{85.34} &48&0.7 \\ \hline
\end{tabular}%
}
\caption{Results of ResNet-50 and ViT When Tested Using Various Unlearning Approaches (in percent accuracy)}
\label{tab:tasks}
\end{figure*}

\section{Methodology}{
We evaluated four paradigms for machine unlearning: 
\begin {enumerate}
\item Fine-tuning: The model is fine-tuned on the remaining dataset, using standard gradient descent techniques. 
\item Pruning + Fine-tuning: First, we apply model pruning to reduce the number of parameters. Then, the pruned model is fine-tuned on the remaining dataset \cite{modelsparsity2024}
\item LoRA: Apply LoRA to selectively modify a subset of the model's parameters.
\item Pruning + LoRA: First prune the model, then add LoRA Adapters and fine-tune.
\end {enumerate}
For our experiments, we employed a ResNet50 and a Vision Transformer (ViT) and trained both on the CIFAR-10 dataset. Our unlearning task focused on removing the influence of the forget class while maintaining performance on the remaining classes. To establish an exact unlearning baseline, we retrain both the models on the remaining dataset for 200 and 90 epochs respectively. We used L2 Pruning to prune 50\% of the specific layers in each model, Convolutional layers were pruned in ResNet50 and linear and attention layers were pruned in ViT. After final finetuning on the remaining dataset for 5/10 epochs, we evaluate the models based on the following metrics:
\begin{itemize}
\item Unlearning accuracy (UA): 1-Acc(Df), where Acc(Df) is the accuracy of the unlearned model on the forget dataset. 
\item Membership inference attack (MIA-Efficacy): Applying the confidence-based MIA predictor to the
unlearned model on the forgetting dataset (Df). A higher MIA-Efficacy implies less information about Df
in the model.
\item Remaining accuracy (RA): This refers to the accuracy of the unlearned model on the retain dataset.
\item Testing accuracy (TA): This refers to the accuracy of the unlearned model on the testing dataset of the remaining classes.
\item Run-time efficiency (RTE): This measures the computation efficiency of the MU method (run time cost).
\item GPU Memory (GPU): This measures the memory requirements of the MU method for a model.
\end{itemize}
}

\section{Results} {
Table \ref{tab:tasks} presents the accuracy metrics for both model under the given five paradigms:
It is observed that all methods achieved perfect or near-perfect Unlearning Accuracy (UA) and Membership Inference Attack (MIA) efficacy, indicating successful removal of the target class information.  For ResNet-50, PruneLoRA outperformed all methods, achieving the highest Remaining (RA) and Testing accuracy (TA), while experiencing near-perfect UA. For the ViT model, PruneLoRA significantly outperformed other methods (except LoRA) in terms of RA and TA. Moreover, while LoRA demonstrated a drastically low UA, PruneLoRA achieved perfect UA. These results suggest that PruneLoRA offers a  balance between effective unlearning, retained model performance, and computational efficiency.
}

\section{Future Scope} {
There is significant potential for further research and experimentation to strengthen and validate our hypothesis. A promising avenue for future research is the application of this method to Large Language Models (LLMs) and Vision-Language Models (VLMs). These models, with their vast parameter spaces, emphasize the need for efficient unlearning techniques. Although computational constraints limited our ability to explore this direction, scaling our approach to these larger models could help develop adaptable and privacy-preserving AI systems.
}

\section{Conclusion} {
This study addresses the challenge of machine unlearning in light of growing privacy regulations and the need for adaptable AI systems. We present a novel approach, PruneLoRA to LoRA to fine-tune sparse models. Our findings highlight the efficacy of LoRA, especially when combined with pruning, in achieving high unlearning performance with minimal computational cost and memory requirements while maintaining general accuracy on remaining classes. These results advances the research in exploring parameter efficient machine approximate unlearning techniques, thus laying the groundwork for applying these methods to complex models such as Large Language Models and Vision-Language Models.
}
\bibliography{aaai25.bib}
\appendix

\section{Appendix}

\section{Experiment Details}{
We trained ResNet-50 and Vision Transformer (ViT) on CIFAR10, using custom implementations. The ResNet-50 model was trained for 200 epochs, and the ViT model was trained for 90 epochs, both on a P100 GPU. They achieved a test accuracy of 95.56\% and 83.77\% respectively. \newline
All further experiments were conducted on a T4 GPU. To allow for meaningful comparison between the various fine-tuning techniques employed, we consistently used the Adam optimizer with a learning rate of $10^{-3}$, along with cross-entropy loss.\\ \newline
In the case of ResNet-50, we applied Structured L2 pruning with a sparsity level of 0.5 across all convolutional layers. Additionally, LoRA was applied to these layers to enable efficient fine-tuning. It is worth noting that future work could explore reducing the number of layers to which LoRA is applied, potentially leading to further computational gains without sacrificing model performance. \newline

For the ViT model, Structured L2 pruning with 0.5 sparsity was applied to the last linear layer and the last attention layer. While initial experiments involved applying LoRA to multiple attention layers, we found that restricting LoRA to the last attention layer yielded the best results. This insight highlights the importance of targeted layer modification in enhancing the model's efficiency. \\

The exact architecture and implementation details for these experiments can be found in the public repository at \url{https://github.com/vlgiitr/LoRA-Unlearn}.
}

\section{Detailed metric settings} {
Details of MIA implementation: MIA is implemented using the prediction confidence-based attack
method. There are mainly two phases during its computation: (1) training phase, and (2)
testing phase. To train an MIA model, we first sample a balanced dataset from the remaining
dataset (Dr) and the test dataset (different from the forgetting dataset Df) to train the MIA predictor.
The learned MIA is then used for MU evaluation in its testing phase. To evaluate the performance
of MU, MIA-Efficacy is obtained by applying the learned MIA predictor to the unlearned model
on the forgetting dataset (Df). Our objective is to find out how many samples in Df can be
correctly predicted as non-training samples by the MIA model.\\ \newline
MIA-Efficacy $=\frac{T N}{\left|\mathcal{D}_{\mathrm{f}}\right|}$ \\ \newline
where TN refers to the true negatives predicted by our MIA predictor, i.e., the number of the
forgetting samples predicted as non-training examples, and $|Df|$ refers to the size of the forgetting
dataset. 
}

\begin{figure}[!t]
    \centering
    \subfloat[Fine-tuned R]{\includegraphics[width=2.5cm,height=2.5cm]{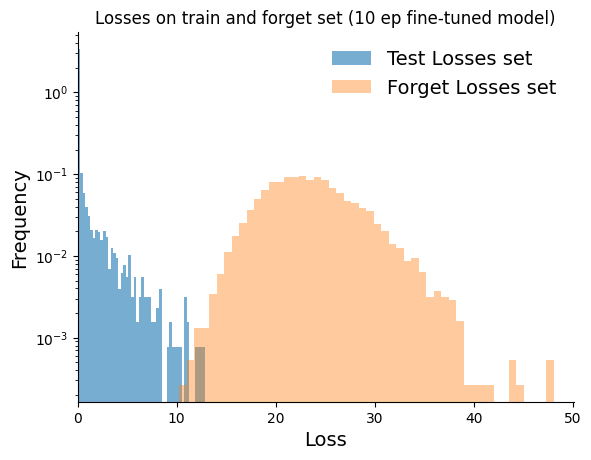}\label{fv1}}%
    \hspace{1mm}
    \subfloat[PruneFT R]{\includegraphics[width=2.5cm,height=2.5cm]{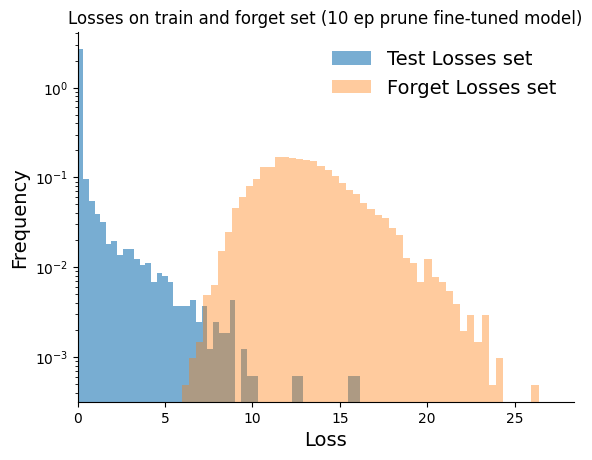}\label{fv2}}%
    \hspace{1mm}
    \subfloat[LoRA R]{\includegraphics[width=2.5cm,height=2.5cm]{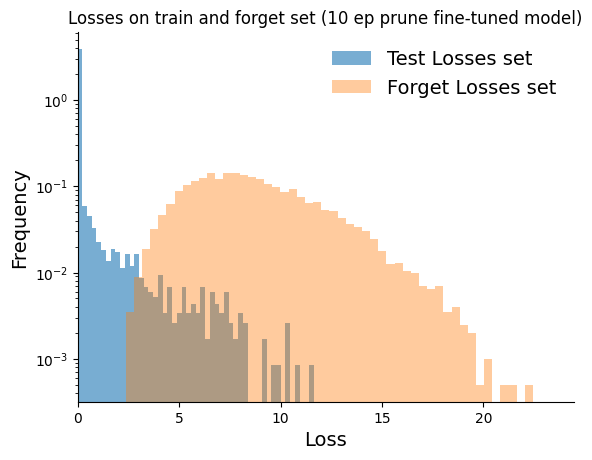}\label{fv3}}%
    \\
    \subfloat[PruneLoRA R]{\includegraphics[width=2.5cm,height=2.5cm]{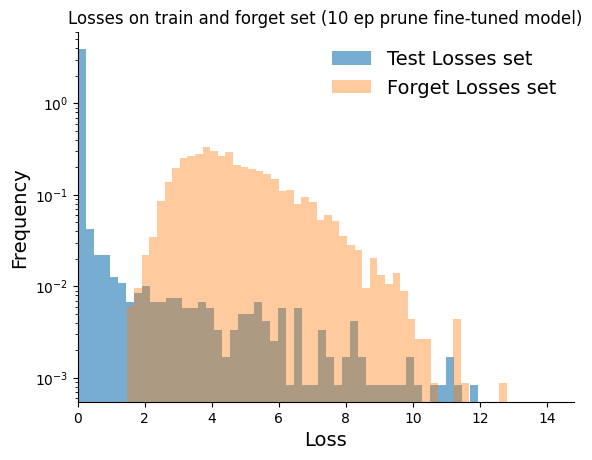}\label{fv4}}%
    \hspace{1mm}
    \subfloat[Fine-tuned V]{\includegraphics[width=2.5cm,height=2.5cm]{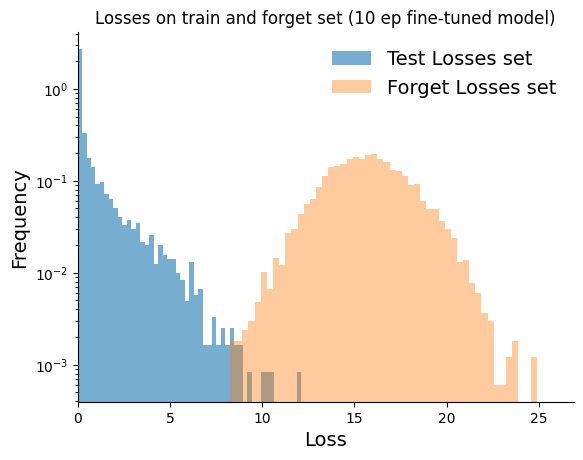}\label{fv5}}%
    \hspace{1mm}
    \subfloat[PruneFT V]{\includegraphics[width=2.5cm,height=2.5cm]{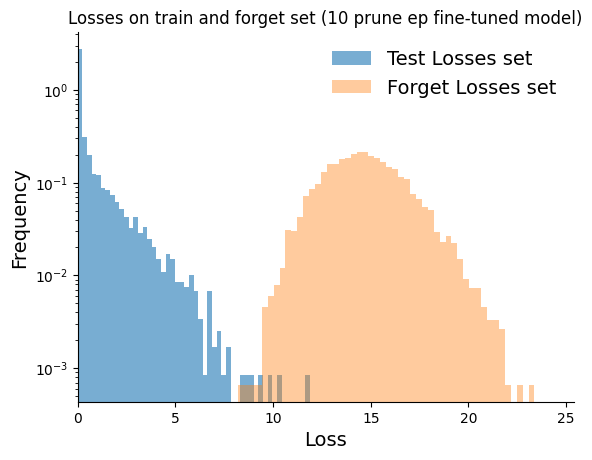}\label{fv6}}%
    \\
    \subfloat[LoRA V]{\includegraphics[width=2.5cm,height=2.5cm]{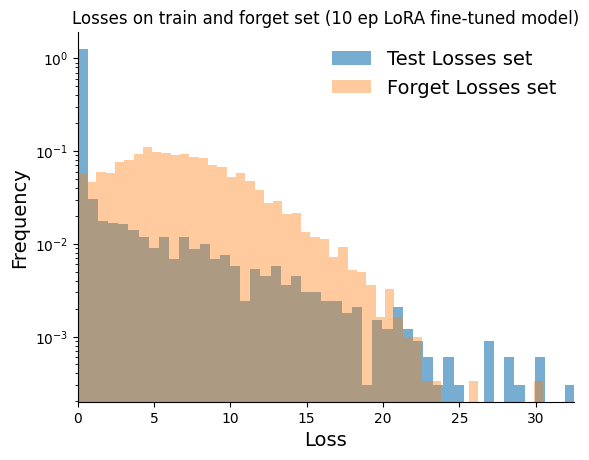}\label{fv7}}%
    \hspace{1mm}
    \subfloat[PruneLoRA V]{\includegraphics[width=2.5cm,height=2.5cm]{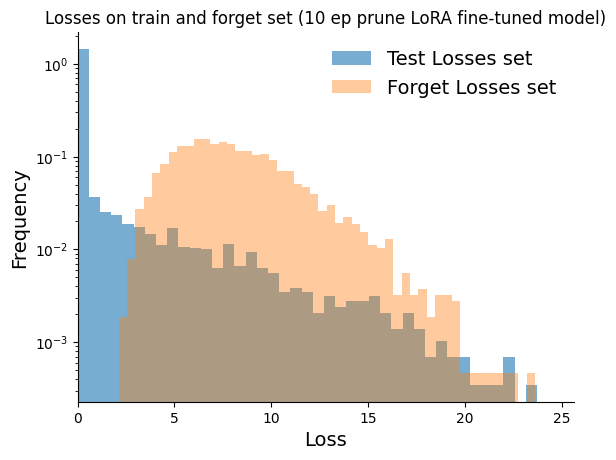}\label{fv8}}%
    \caption{Visual Comparison of Cross-Entropy Losses of Test and Forget Set (R: ResNet-50 V: ViT) Forget Losses must be higher than Test losses for  Unlearning Accuracy}
    \label{fig:Validation}
\end{figure}

\section{Future Scope} {
Due to lack of computational resources and funding, we were only able to perform a limited number of experiments, without such constraints, another approach is to include layer-specific adaptation strategies where different layers or components of the model are subject to distinct unlearning approaches after studying optimal unlearning strategies for the respective layers. Models can also be studied under continual learning contexts, where models repeatedly learn and unlearn information over time. 
}
\end{document}